\documentclass{article}

\usepackage{arxiv}

\usepackage[utf8]{inputenc} % allow utf-8 input
\usepackage[T1]{fontenc}    % use 8-bit T1 fonts
\usepackage{hyperref}       % hyperlinks
\usepackage{url}            % simple URL typesetting
\usepackage{booktabs}       % professional-quality tables
\usepackage{amsfonts}       % blackboard math symbols
\usepackage{nicefrac}       % compact symbols for 1/2, etc.
\usepackage{microtype}      % microtypography
\usepackage{lipsum}		% Can be removed after putting your text content
\usepackage{graphicx}
\usepackage[numbers]{natbib}  % [numbers] untuk citation dengan angka dalam kurung siku
\usepackage{doi}

\title{Classification of Public Opinion on the Free Nutritional Meal Program on YouTube Media Using the LSTM Method}

\author{
	\textbf{Berliana Enda Putri} \\
	Department of Data Science \\
	Institut Teknologi Sumatera \\
	Lampung Selatan, 35365, Indonesia \\
	\texttt{berliana.122450065@student.itera.ac.id} \\
	\And
	\textbf{Lisa Diani Amelia} \\
	Department of Data Science \\
	Institut Teknologi Sumatera \\
	Lampung Selatan, 35365, Indonesia \\
	\texttt{lisa.122450021@student.itera.ac.id} \\
	\And
	\textbf{Muhammad Zaky Zaiddan} \\
	Department of Data Science \\
	Institut Teknologi Sumatera \\
	Lampung Selatan, 35365, Indonesia \\
	\texttt{zaky.122450119@student.itera.ac.id} \\
    \And
    \textbf{Luluk Muthoharoh, M.Si} \\
    Department of Data Science \\
    Institut Teknologi Sumatera \\
    Lampung Selatan, 35365, Indonesia \\
    \texttt{luluk.muthoharoh@sd.itera.ac.id} \\
    \And
    \textbf{Ardika Satria, M.Si} \\
    Department of Data Science \\
    Institut Teknologi Sumatera \\
    Lampung Selatan, 35365, Indonesia \\
    \texttt{ardika.satria@sd.itera.ac.id} \\  
    \And
    \textbf{Martin Clinton Tosima} \\
    \textbf{Manullang, Ph.D.} \\
    Department of Informatics \\
    Institut Teknologi Sumatera \\
    Lampung Selatan, 35365, Indonesia \\
    \texttt{martin.manullang@if.itera.ac.id} \\
}

\renewcommand{\headeright}{}
\renewcommand{\undertitle}{}
\renewcommand{\shorttitle}{}

\hypersetup{
pdftitle={Classification of Public Opinion on the Free Nutritional Meal Program on YouTube Media Using the LSTM Method},
pdfsubject={Sentiment Analysis, LSTM, YouTube Comments},
pdfauthor={Berliana Enda Putri, Lisa Diani Amelia, Muhammad Zaky Zaiddan, Martin Clinton Tosima Manullang},
pdfkeywords={LSTM, Sentiment Analysis, YouTube, Free Nutritional Meal, Classification},
}

\begin{document}
\maketitle

\begin{abstract}
Public opinion towards the Free Nutritious Meal Program (MBG) on YouTube social media reflects the diverse responses of the community. This study implements the Long Short-Term Memory (LSTM) method to classify sentiments from 7,733 YouTube comments. The results show that the LSTM model achieves an accuracy of 89\%, with excellent performance on negative sentiment (f1-score 0.94) but limited performance on positive sentiment (f1-score 0.55) due to the dominance of negative data (87.7\%). These findings confirm the effectiveness of LSTM in sentiment analysis of Indonesian text while highlighting the challenge of data imbalance. This research provides practical contributions for social media-based public policy evaluation.
\end{abstract}


\section{Introduction}
The development of digital technology and social media has transformed the way people express their opinions on public policies. Platforms such as \textit{YouTube} have become open interaction spaces that allow users to provide comments, support, or criticism directly on government programs. These comments generate large volumes of textual data that can be utilized to understand public perception more quickly and efficiently compared to conventional survey methods. Therefore, public opinion analysis through social media has become an important approach in data-driven decision making \cite{liu2012}.

One policy that has attracted considerable public attention in Indonesia is the Free Nutritious Meal Program (MBG). This program aims to improve the nutritional quality of the community, particularly school children, and to support long-term human resource development. However, the implementation of public policies often elicits various responses from the public, including positive, negative, and neutral opinions. Thus, an approach capable of systematically identifying public sentiment is needed to determine the level of public acceptance of the program \cite{pak2010}.

In the field of Natural Language Processing (NLP), sentiment analysis is a technique used to identify the polarity of opinions in a text into specific categories, such as positive, negative, or neutral. This method has been widely used to evaluate public responses to various objects, including products, services, and government policies. By utilizing comment data from social media, sentiment analysis can provide a general overview of public perception in real-time \cite{medhat2014}.

One of the deep learning methods widely used in text classification is Long Short-Term Memory (LSTM). LSTM is a development of Recurrent Neural Network (RNN) designed to address the vanishing gradient problem and to capture long-term dependencies between words in a text sequence. This capability makes LSTM highly suitable for sentiment classification tasks, as sentence context and word order significantly influence the meaning of the expressed opinion \cite{hochreiter1997}.

Based on this background, this study aims to classify public opinion on the Free Nutritious Meal Program on YouTube media using the LSTM method. The results of this study are expected to provide an overview of public perception of the policy and serve as evaluation material for the government in improving the effectiveness of the MBG program implementation.

\section{Related Work}

\subsection{Sentiment Analysis on Social Media}
Sentiment analysis is a branch of Natural Language Processing (NLP) that aims to identify opinions, emotions, or attitudes of individuals towards a particular topic based on textual data. Sentiment analysis generally classifies text into positive, negative, and neutral categories \cite{liu2012}. This technique has been widely used to understand public perception of products, services, public figures, and government policies.

The development of social media has made digital platforms a primary data source for sentiment analysis. Social media platforms such as Twitter, Facebook, Instagram, and YouTube allow users to express their opinions openly through comments or posts. This data is real-time, large in volume, and reflects public opinion directly, making it highly valuable for data-driven decision making \cite{pak2010}.

Various previous studies have utilized social media for sentiment analysis. Medhat et al. conducted a comprehensive survey of sentiment analysis methods and demonstrated that text classification approaches are effective for extracting public opinions from online data \cite{medhat2014}. Furthermore, Jose and Simritha applied LSTM to social media for sentiment analysis and topic extraction, demonstrating the model's ability to understand user emotions \cite{jose2024}.

In the context of public policy, sentiment analysis of public comments can serve as a fast and efficient evaluation tool. Therefore, this study uses YouTube comments as a data source to understand public perception of the Free Nutritious Meal Program (MBG).

\subsection{Text Classification with LSTM}
Text classification is the process of automatically grouping documents or sentences into specific categories using computational algorithms. In sentiment analysis, text classification is used to determine whether a comment contains positive, negative, or neutral sentiment.

Traditional classification methods such as Naive Bayes, Logistic Regression, and Support Vector Machine (SVM) have been widely used in text classification tasks. However, these methods generally rely on feature representations such as Bag of Words or TF-IDF, which are not fully capable of capturing the contextual order of words in a sentence.

With the advancement of deep learning, Long Short-Term Memory (LSTM) has become a popular approach in text classification. LSTM is a development of Recurrent Neural Network (RNN) designed to address the vanishing gradient problem in sequential data \cite{hochreiter1997}. The LSTM architecture features a memory cell, input gate, forget gate, and output gate that allow the model to retain important information from previous words.

Several studies have demonstrated the strong performance of LSTM in sentiment analysis tasks. Bargavi and Rekha stated that LSTM is more effective than CNN and RNN in text-based sentiment classification \cite{bargavi2021}. Another study by Waluyo and Juwono used LSTM with TensorFlow to classify negative comments on social media and obtained satisfactory results \cite{waluyo2023}. Ahmed et al. also compared RNN and LSTM on airline tweet data and showed that LSTM has more stable performance on sequential data \cite{ahmed2025}.

The main advantage of LSTM is its ability to understand relationships between words in a sentence sequence. This is crucial in sentiment analysis because the meaning of a sentence is often influenced by the context of previous words. For example, the phrase "not good" has a negative meaning even though it contains the word "good," which has a positive value.

In this study, LSTM is used as the primary model to classify public opinion on the Free Nutritious Meal Program (MBG). The model’s performance is then compared with other methods such as SVM, Logistic Regression, and Naive Bayes to determine the effectiveness of LSTM in handling Indonesian-language YouTube comment data. The comparison between deep learning and traditional machine learning is essential to validate model robustness in various linguistic contexts \cite{onan2016}. Furthermore, previous research on Indonesian social media sentiment has shown that while traditional methods like SVM are reliable, they often require extensive feature engineering compared to neural network approaches \cite{alfina2017}.

\section{Methodology}
This study employs a quantitative approach using Natural Language Processing (NLP) methods to classify public opinion on the Free Nutritious Meal Program (MBG) based on comments on the YouTube platform. The research stages include data collection, text preprocessing, data representation, modeling, and model performance evaluation.

\subsection{Data Collection}
The dataset was obtained through a scraping process of comments from two YouTube channels discussing the Free Nutritious Meal Program (MBG). A total of 7,733 comments were successfully collected, with details as follows:

\begin{table}[h]
\centering
\caption{Dataset Summary}
\begin{tabular}{lc}
\toprule
Source & Number of Comments \\
\midrule
YouTube Channel 1 & 3,342 \\
YouTube Channel 2 & 4,391 \\
\midrule
Total & 7,733 \\
\bottomrule
\end{tabular}
\label{tab:dataset}
\end{table}

The data obtained are unstructured text containing informal language, abbreviations, and various public opinions.

\subsection{Data Preprocessing}
The preprocessing stage aims to clean and prepare the raw data for the classification process. The preprocessing steps performed are as follows:

\begin{enumerate}
\item \textbf{Case Folding}: Change all uppercase text to lowercase.Then, remove unnecessary characters such as URLs, mentions, hashtags, numbers,punctuation, and non-alphabetic characters.
\item \textbf{Text Cleaning}: Remove URLs, mentions, hashtags, numbers, punctuation, and special characters.
\item \textbf{Normalization}: Convert non-standard or slang words into standard forms.
\item \textbf{Tokenization}: Break sentences into words or tokens.
\item \textbf{Stopword Removal}: Remove common words that have no significant meaning using NLTK's stopword list, such as abbreviations and conjunctions.
\item \textbf{Stemming}: Convert affixed words into root words using the Sastrawi library.
\end{enumerate}

\subsection{Feature Representation}
After preprocessing, the text data is converted into numerical form using a Tokenizer. Each word is converted into an integer index. Subsequently, padding is performed so that all text sequences have the same length, enabling them to be processed by the LSTM model.

\subsection{Classification Models}
This study uses several classification models to compare performance, including:

\begin{itemize}
\item Long Short-Term Memory (LSTM)
\item Support Vector Machine (SVM)
\item Logistic Regression
\item Naive Bayes
\end{itemize}

LSTM is used as the main model because it is capable of capturing sequential relationships between words in the text.

\subsection{Sentiment Analysis}

Sentiment analysis was performed using the Long Short-Term Memory (LSTM) model after the data had undergone preprocessing, tokenization, and padding. The model was trained using the training data and validated with the validation data to evaluate its performance. The architecture of the LSTM model used in this study consists of several main layers:

\begin{itemize}
    \item \textbf{Embedding Layer}: Converts words into 128-dimensional vectors. This layer uses a padding index of 0 to handle sequences of varying lengths.
    \item \textbf{LSTM Layer}: Processes word sequences sequentially with a hidden dimension of 128 and a dropout rate of 0.3. This layer captures long-term dependencies between words in a sentence.
    \item \textbf{Dropout Layer}: Prevents overfitting by randomly dropping neurons during training with a dropout rate of 0.5.
    \item \textbf{Fully Connected Layer}: Connects the extracted features to the classification layer with 2 output neurons (positive and negative).
    \item \textbf{Output Layer}: Uses the Softmax activation function to classify sentiments into two categories: positive and negative.
\end{itemize}

\begin{table}[htbp]
\centering
\caption{LSTM Hyperparameters}
\begin{tabular}{|l|c|}
\hline
\textbf{Hyperparameter} & \textbf{Value} \\
\hline
Vocabulary Size & 16,378 \\
Embedding Dimension & 128 \\
Hidden Dimension & 128 \\
Total Parameters & 2,228,738 \\
Batch Size & 16 \\
Learning Rate & 0.0005 \\
Epochs & 20 \\
Dropout (LSTM) & 0.3 \\
Dropout (FC) & 0.5 \\
Output Activation Function & Softmax \\
Optimizer & Adam \\
Loss Function & CrossEntropyLoss \\
\hline
\end{tabular}
\label{tab:hyperparameter_lstm}
\end{table}

\subsection{Evaluation Method}
The final evaluation stage involves assessing the sentiment classification results using a confusion matrix and a classification report. Four evaluation metrics are used, namely:

\begin{itemize}
    \item \textbf{Accuracy}: The proportion of correct predictions out of all predictions made.
    \begin{equation}
        Accuracy = \frac{TP + TN}{TP + TN + FP + FN}
    \end{equation}
    
    \item \textbf{Recall}: The model's success rate in finding all positive data.
    \begin{equation}
        Recall = \frac{TP}{TP + FN}
    \end{equation}
    
    \item \textbf{F1-Score}: The harmonic mean of precision and recall.
    \begin{equation}
        F1 = 2 \times \frac{Precision \times Recall}{Precision + Recall}
    \end{equation}
    
    \item \textbf{Precision}: The accuracy level of positive predictions made by the model.
    \begin{equation}
        Precision = \frac{TP}{TP + FP}
    \end{equation}
\end{itemize}

The following are the components of the confusion matrix:

\begin{itemize}
    \item \textbf{TP (True Positive)}: Positive data predicted as positive by the model.
    \item \textbf{FP (False Positive)}: Negative data predicted as positive by the model.
    \item \textbf{TN (True Negative)}: Negative data predicted as negative by the model.
    \item \textbf{FN (False Negative)}: Positive data predicted as negative by the model.
\end{itemize}

Furthermore, Table 2 presents an overview of the model's performance in classifying data to show the distribution of predictions between classes.

\begin{table}[htbp]
\centering
\caption{Confusion Matrix}
\begin{tabular}{|l|c|c|}
\hline
 & \textbf{Predicted Positive} & \textbf{Predicted Negative} \\
\hline
\textbf{Actual Positive} & TP (True Positive) & FN (False Negative) \\
\hline
\textbf{Actual Negative} & FP (False Positive) & TN (True Negative) \\
\hline
\end{tabular}
\label{tab:confusion_matrix}
\end{table}

\section{Results and Discussion}

\subsection{Data Distribution}
At this stage, the distribution of comment sentiments related to the MBG program is analyzed. This study has 87.7\% negative labels and 12.3\% positive labels. From the labeling results, data with negative sentiment amounted to 5,629 and positive sentiment amounted to 790. The distribution of sentiment analysis results is presented in
Figure 1. 
\begin{center}
    \includegraphics[width=0.5\linewidth]{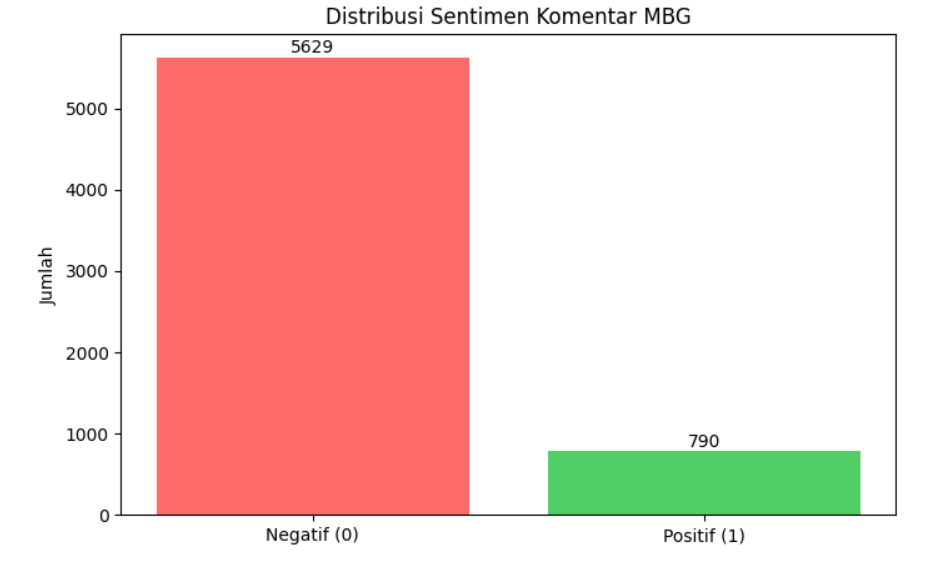}
    \par
    \textbf{Figure 1:} Distribution of MBG Comment Sentiments
\end{center}

\subsection{LSTM Model Results}
The Long Short-Term Memory (LSTM) model was trained using data that had undergone preprocessing with a total vocabulary of 16,378 unique tokens. The training process was conducted for 20 epochs using the Adam optimizer with a learning rate of 0.0005 and categorical crossentropy loss function. The model has a total of 2,228,738 parameters. Based on the training results, the model showed reasonably good performance with a training accuracy of 98.26\% and a validation accuracy of 88.99\%. The loss values tended to decrease from 0.68 at the initial epoch to 0.1225 at the 20th epoch, indicating that the model was able to learn the data patterns well. The visualization of the training process, including the training and validation loss as well as the training and validation accuracy over 20 epochs, is presented in Figure 2. This figure illustrates how the model's performance improved progressively throughout the training phase.

\begin{center}
    \includegraphics[width=0.9\linewidth]{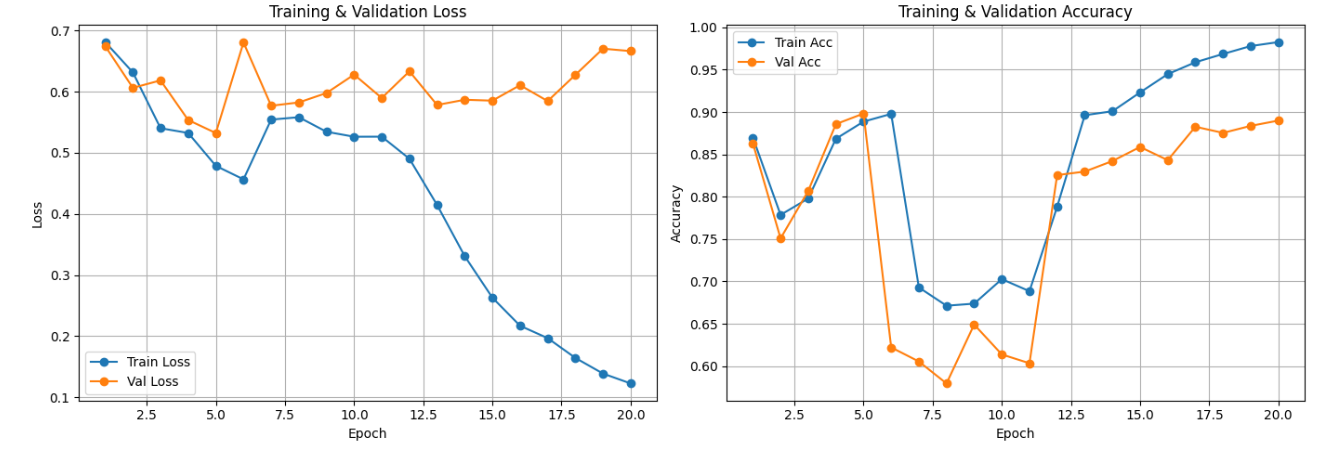}
    \par
    \textbf{Figure 2:} Training \& Validation Loss and Accuracy of the LSTM Model
\end{center}

\subsection{Model Evaluation}
Evaluation of the model on 963 test samples yielded an accuracy of 89\%. For the negative class, the model achieved a precision of 0.94, recall of 0.93, and f1-score of 0.94. Meanwhile, for the positive class, the model achieved a precision of 0.54, recall of 0.57, and f1-score of 0.55. These results indicate that the LSTM model is quite effective in performing sentiment classification on YouTube comment data, especially for the negative class. The evaluation results are presented in Table 1.

\begin{table}[htbp]
\centering
\caption{LSTM Model Evaluation Results}
\begin{tabular}{lcccc}
\toprule
Sentiment & Precision & Recall & F1-Score & Support \\
\midrule
Negative & 0.94 & 0.93 & 0.94 & 845 \\
Positive & 0.54 & 0.57 & 0.55 & 118 \\
\midrule
Accuracy & \multicolumn{4}{c}{0.89} \\
Macro Avg & 0.74 & 0.75 & 0.74 & 963 \\
Weighted Avg & 0.89 & 0.89 & 0.89 & 963 \\
\bottomrule
\end{tabular}
\label{tab:evaluasi}
\end{table}

The confusion matrix in Figure 3 provides a detailed visualization of the model's classification performance across both sentiment classes.

\begin{center}
    \includegraphics[width=0.7\linewidth]{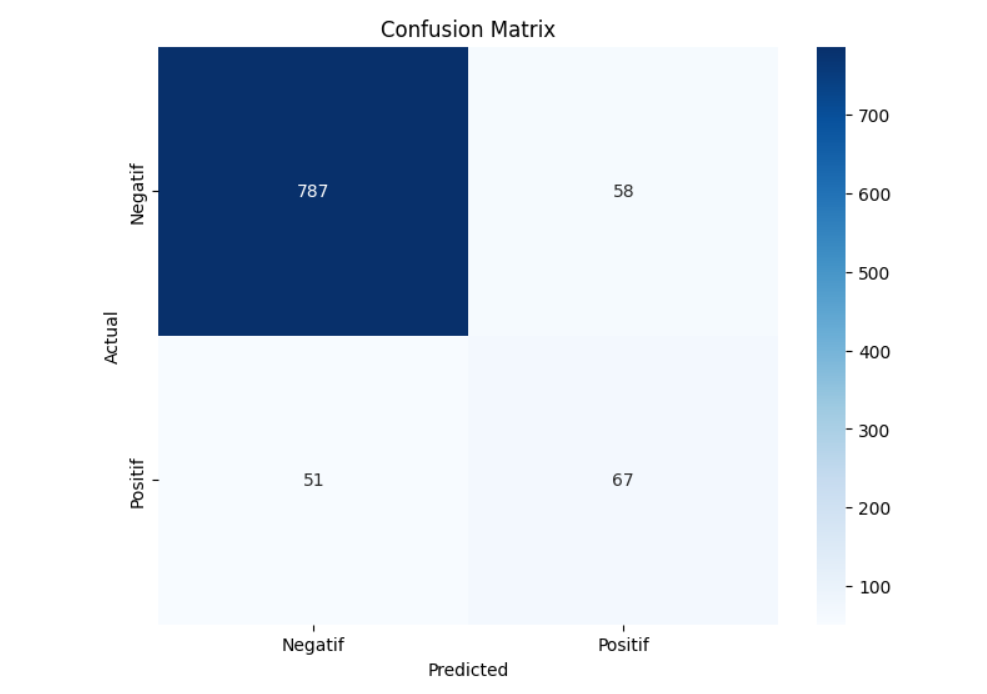}
    \par
    \textbf{Figure 3:} Confusion Matrix
\end{center}

Based on the confusion matrix shown in Figure 3, the model correctly classified 787 negative comments as negative (True Negatives) and 67 positive comments as positive (True Positives). However, the model misclassified 51 positive comments as negative (False Negatives), indicating that a substantial number of positive sentiments were not recognized by the model. Additionally, 58 negative comments were incorrectly predicted as positive (False Positives). The high number of False Negatives (51) confirms the earlier finding that the model struggles to identify positive sentiment, primarily due to the imbalanced dataset where negative samples dominate. This suggests that the model tends to bias its predictions toward the negative class. Improving data balance through techniques such as SMOTE or collecting more positive samples could help reduce False Negatives and enhance overall classification performance.

\section{Conclusion}
This study successfully implemented the Long Short-Term Memory (LSTM) method for classifying public opinion on the Free Nutritious Meal Program (MBG) using YouTube comment data. The model achieved an accuracy of 89\% on the test dataset, demonstrating its effectiveness in sentiment analysis tasks for Indonesian-language text. Specifically, the model exhibited excellent performance in identifying negative sentiment with an F1-score of 0.94, indicating its strength in capturing critical public responses. However, the model's performance on positive sentiment remained limited with an F1-score of 0.55, primarily due to the imbalanced dataset dominated by negative comments (87.7\%). This finding highlights the importance of data balance in deep learning-based classification tasks. For future research, it is recommended to apply data balancing techniques such as SMOTE or class weighting, increase the collection of positive sentiment data, and explore advanced architectures like Bi-LSTM or CNN-LSTM to further improve classification performance, particularly for minority classes.

\end{document}